\documentclass[11pt]{article}

\usepackage[preprint]{acl}

\usepackage{times}
\usepackage[T1]{fontenc}
\usepackage[utf8]{inputenc}
\usepackage{microtype}
\usepackage{inconsolata}
\usepackage{fvextra}
\usepackage{needspace}
\usepackage{graphicx}
\usepackage{caption}
\usepackage{float}
\usepackage{afterpage}
\usepackage{booktabs}
\usepackage{multirow}
\usepackage{tabularx}
\usepackage{amsmath}
\usepackage{xcolor}
\usepackage{colortbl}

\title{Abstract What Varies, Reuse Longer Patterns: Efficient Offline Chain-of-Thought Compression}

\author{%
  Shibo Wang\textsuperscript{1}\quad Yuzhe He\textsuperscript{2}\quad
  Yifeng Zhao\textsuperscript{1}\quad Yunjiao Zhou\textsuperscript{1}\\
  \bfseries Zepeng Ning\textsuperscript{3}\quad Hanzhang Zhou\textsuperscript{1}\quad
  Kezhi Mao\textsuperscript{1}\quad Junlang Qian\textsuperscript{1}\\[3pt]
  {\normalfont\small\textsuperscript{1}Nanyang Technological University\qquad
   \textsuperscript{2}City University of Hong Kong}\\
  {\normalfont\small\textsuperscript{3}Nankai University}%
}

\begin{document}
\raggedbottom
\maketitle

\begin{abstract}
Offline chain-of-thought (CoT) compression creates shorter supervision for
efficient reasoning, but per-example compressors repeatedly process similar
expressions across traces, wasting compressor budget. Methods that reuse identical text across
examples miss longer shared expressions when names, numbers, or formulas
change from one question to another. We introduce CRAFT, which abstracts
problem-specific details into slots to uncover partially lexicalized
expression templates (PLETs) and clusters wording variants into families.
Compressing each family representative once yields shared deletion or
rewriting edits, which are applied with each occurrence's own details to
construct compact, ordinary-text training targets. Compared with
per-example rewriting, CRAFT Rewrite reduces weighted compressor budgets
by a median 90.7\% with a median accuracy loss of 1.21 percentage points,
while generating shorter reasoning. Against cross-sample encoding, it
shortens reasoning across all eight model--dataset settings by 25.3\%
on average and improves accuracy in five.
\end{abstract}

\section{Introduction}
\label{sec:introduction}

\suppressfloats[t]
\begin{figure}[t]
  \centering
  \includegraphics[width=\linewidth]{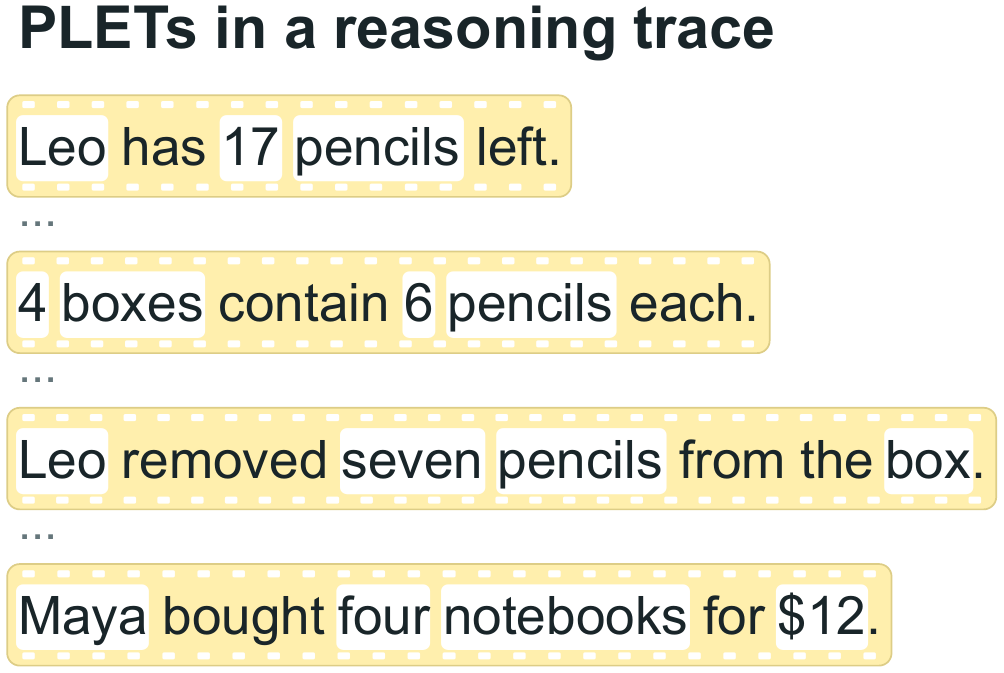}
  \caption{PLETs combine reusable wording with problem-specific entities,
  values, and formulas in a reasoning trace.}
  \label{fig:pareo-motivation}
  \vspace{-1em}
\end{figure}

Chain-of-thought (CoT) improves reasoning through additional test-time
computation, but autoregressive generation makes long traces costly,
especially for larger models
\citep{wei-etal-2022-chain,snell-etal-2025-test-time-compute,han-etal-2025-token,ma-etal-2025-cot-valve}.
Offline CoT compression methods often shorten model-generated traces
through per-example deletion or rewriting, then use the compressed
traces for supervised fine-tuning (SFT),
teaching models to generate shorter reasoning directly
\citep{xia-etal-2025-tokenskip,kang-etal-2024-c3ot,yan-etal-2025-macc}.
However, per-example compression incurs substantial generation costs
when using a powerful teacher model as the compressor.

\afterpage{\afterpage{%
\begin{figure}[!t]
  \centering
  \includegraphics[width=\linewidth]{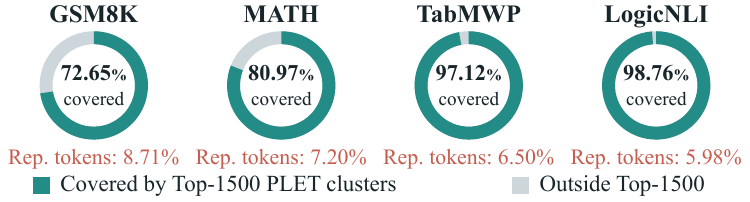}
  \caption{PLET coverage and reuse across reasoning corpora.}
  \label{fig:plet-coverage}
  \vspace{-1em}
\end{figure}
}}

Per-example compression removes redundancy within each trace but repeats
work across traces. Within a domain, different questions often elicit
the same wording for relating quantities, explaining facts, or describing
solution steps; their entities and calculations change while much of the
surrounding expression recurs. Compressing each question--trace pair
independently sends these expressions to the compressor repeatedly and,
for rewriting, regenerates similar shortened wording. Sharing compression
decisions across similar expressions in different traces could reduce
the compressor budget.

The challenge is that reusable expressions need not be identical.
Methods based on exact sequences, including dictionary compression and
supertoken vocabularies \citep{ziv-lempel-1977-universal,zhao-etal-2026-shorthand},
face two sources of variation. First, problem-specific names, numbers,
and formulas make sentences built from the same template look different.
Abstracting these details into slots reveals Partially Lexicalized
Expression Templates (PLETs): shared wording with variable content
(Figure~\ref{fig:pareo-motivation}).\footnote{The term
follows the combination of recurring wording and variable content in
formulaic language
\citep{fillmore-etal-1988-regularity,erman-warren-2000-idiom,constant-etal-2017-survey}.}
Second, minor paraphrases can leave templates different in wording
even when they convey the same meaning. Clustering these variants into
families extends reuse beyond identical templates. Together, slot
abstraction and clustering enable shared compression across differences
in both problem-specific content and wording.

We introduce \textbf{CRAFT} (\textbf{C}ompression through
\textbf{R}eusable \textbf{A}bstractions and \textbf{F}requent
\textbf{T}emplates). CRAFT abstracts problem-specific details into PLETs,
clusters similar PLETs, compresses each cluster's representative through
deletion or rewriting, and restores each occurrence's details to produce
compressed CoTs for SFT.
Our evaluation spans Qwen2.5 (3B/7B/14B), LLaMA-3.1-8B, and four
benchmarks: GSM8K, MATH-500, TabMWP, and LogicNLI.
Against per-example methods, we compare length--compressor-budget
tradeoffs under accuracy matching and accuracy--compressor-budget
tradeoffs under length matching; against low-resource and cross-sample
baselines, we compare accuracy--length tradeoffs.
Against per-example rewriting, CRAFT Rewrite reduces weighted compressor
budgets by a median 90.7\% with a median accuracy loss of 1.21 percentage
points, while generating shorter reasoning (Figure~\ref{fig:matched-budget}).
Against cross-sample encoding, it shortens reasoning by 25.3\% on average
across eight settings and improves accuracy in five.
Compared with each zero-cost baseline's highest-accuracy evaluated
setting, it improves both accuracy and length in seven of eight settings
for truncation and four of eight for length-controlled prompting
(Table~\ref{tab:low-resource-comparison}).

Our contributions are threefold:
\begin{itemize}
  \item We identify reusable structure hidden by problem-specific details
  and wording variations, exposing compression opportunities beyond
  identical text.
  \item We introduce CRAFT, which combines slot abstraction and template
  clustering to share compression decisions across traces and construct
  compact CoT supervision.
  \item We demonstrate substantial compressor-budget savings over
  per-example methods and stronger accuracy--length tradeoffs against
  cross-sample encoding and zero-cost baselines.
\end{itemize}

\section{Related Work}
\label{sec:related-work}

\paragraph{Offline CoT compression for SFT.}
Compact SFT targets use token pruning, including
TokenSkip's LLMLingua-2 scoring
\citep{pan-etal-2024-llmlingua,xia-etal-2025-tokenskip}, or larger-unit
selection \citep{li-etal-2026-step-entropy,ma-etal-2026-reasoning}.
Other methods use rewriting, iterative refinement, or multi-stage
compression
\citep{kang-etal-2024-c3ot,yan-etal-2025-macc,fan-etal-2026-ctrlcot,wang-etal-2025-r1-compress}.
Other approaches control reasoning length through tuning
\citep{ma-etal-2025-cot-valve} or combine compressed-target SFT with
reinforcement learning \citep{tang-etal-2026-extra-cot}.
CRAFT shares template-family compression policies across recurring
expressions in different traces.

\paragraph{Reusable reasoning templates.}
Prior work selects demonstrations \citep{zhang-etal-2023-auto-cot},
composes or reuses high-level thought structures
\citep{yang-etal-2024-buffer,zhou-etal-2024-self-discover}, or retrieves
and organizes reasoning steps and paths
\citep{ahmed-etal-2026-retrieval,yang-etal-2026-drpc} to guide problem
solving or rationale generation.
CRAFT reuses editing policies over local expression templates in
already-generated traces, with each occurrence supplying its own slot
bindings.

\paragraph{Formulaic expressions and repetition-based compression.}
Corpus research documents widespread prefabricated expressions
in English \citep{erman-warren-2000-idiom}, while
multiword-expression research highlights variability within recurring
expressions \citep{constant-etal-2017-survey}.
Fixed lexical material with open positions
\citep{fillmore-etal-1988-regularity} motivates our text-level PLET
representation of stable wording and variable slots.
Lempel-Ziv compression encodes repeated substrings through references to
earlier input \citep{ziv-lempel-1977-universal}, whereas byte-pair
encoding repeatedly replaces frequent adjacent symbol pairs
\citep{gage-1994-compression}; both support exact reconstruction.
Cross-word vocabulary compression similarly assigns new tokens to fixed
strings \citep{zhao-etal-2026-shorthand}.
CRAFT shares potentially lossy deletion or rewriting
policies over templates through
occurrence-local alignment and slot binding under the existing tokenizer
and vocabulary.

\section{Method}
\label{sec:method}

\begin{figure*}[t]
  \centering
  \includegraphics[width=\textwidth]{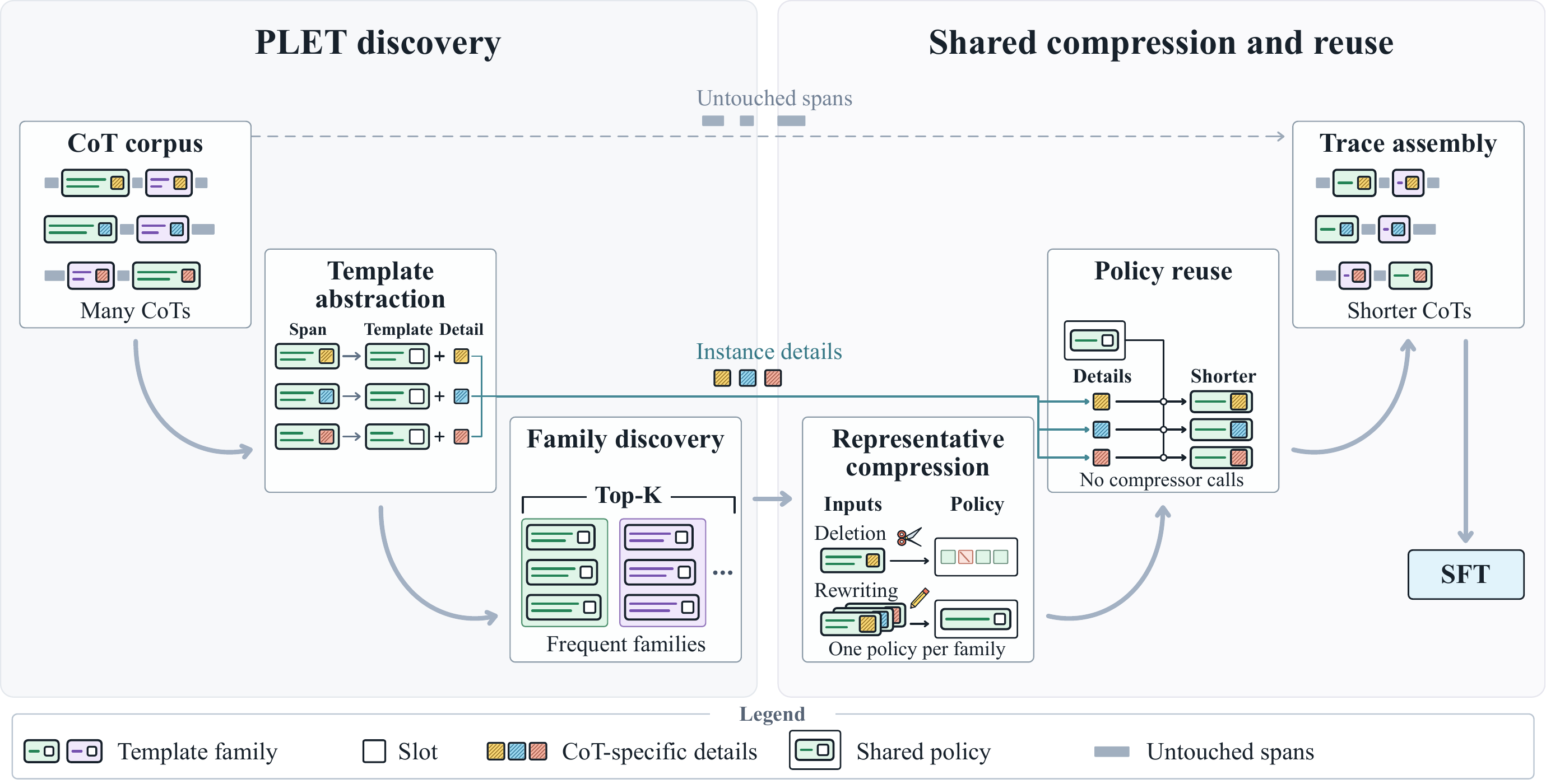}
  \caption{Overview of PLET discovery, shared compression, and reuse.
  Local spans are abstracted and grouped into families, prioritized by
  source-token mass. Shared policies are applied using occurrence content
  and bindings, and compact traces are assembled as SFT targets while
  preserving source order and unselected content.}
  \label{fig:method-overview}
\end{figure*}

\subsection{Problem Formulation}
\label{sec:method-setting}

Let $\mathcal D=\{(q_i,c_i,y_i)\}_{i=1}^{N}$ be a corpus of questions
$q_i$, chain-of-thought (CoT) traces $c_i$, and final answers $y_i$.
Each trace is an ordered sequence of non-overlapping local spans, or
units, $\mathcal U_i=(u_{i,1},\ldots,u_{i,M_i})$.
Offline CoT compression seeks shorter traces $\hat c_i$ that retain useful
reasoning supervision, yielding compact question--response pairs for
supervised fine-tuning (SFT).  We denote the length of a text $x$ under the
target model's tokenizer by $\ell(x)$.
Let $\mathcal O$ be a deletion \citep{xia-etal-2025-tokenskip} or rewriting
\citep{kang-etal-2024-c3ot,yan-etal-2025-macc} operator with setting $\eta$.
Instead of compressing each occurrence separately, CRAFT groups candidate
spans into template families and reuses one policy within each family.

\subsection{Template Abstraction}
\label{sec:method-abstraction}

CRAFT segments each trace at common
semantic boundaries, such as sentences and lists, to obtain
$\mathcal U_i=\operatorname{Seg}(c_i)$. We enumerate all valid windows of one to three
consecutive units as candidates $s_{i,n}$, retaining their source intervals
for trace assembly. Multi-unit windows enable abstraction and clustering
to discover PLETs of varying lengths across unit boundaries. Question-guided
abstraction separates problem-specific details from reusable wording.

Let $C(q_i)$ contain the non-stopword
noun, proper-noun, and adjective lemmas in question $q_i$, and let
$\operatorname{df}(w)$ count questions containing lemma $w$, once per question.
For a frequency proportion $\rho_{\mathrm{df}}$, set
$\tau_{\mathrm{df}}=\max\{2,\lceil\rho_{\mathrm{df}}N\rceil\}$ and select
\begin{equation}
  V_i=\{w\in C(q_i):\operatorname{df}(w)\leq\tau_{\mathrm{df}}\}
      ,
  \label{eq:pareo-question-vocabulary}
\end{equation}
Words in $c_i$ whose lemmas belong to $V_i$ become entity slots.
Abstracting each candidate as a whole yields
$A(q_i,s_{i,n})=(a_{i,n},B_{i,n})$: the normalized PLET $a_{i,n}$ retains
wording and slot structure for comparison, while
$B_{i,n}=\{\sigma_k\mapsto v_k\}_{k=1}^{H_{i,n}}$ preserves occurrence-specific
contents $v_k$ for its $H_{i,n}$ local slots $\sigma_k$.

\subsection{High-Frequency Template Discovery}
\label{sec:method-discovery}

PLETs can still differ in wording after abstraction. We first aggregate
identical forms, retaining their occurrences and bindings, then cluster
distinct forms into families. For forms $a_p,a_t$, lexical similarity
$L_{pt}$ captures wording overlap through cosine similarity of word-unigram
and bigram TF--IDF vectors; semantic similarity $E_{pt}$ uses frozen sentence
embeddings to capture similar meanings. Edit compatibility compares
ordered slot-type sequences $z_p,z_t$:
\begin{equation}
  C_{pt}=\frac{\operatorname{LCS}(z_p,z_t)}{\max\{|z_p|,|z_t|\}},
  \label{eq:pareo-slot-similarity}
\end{equation}
where $\operatorname{LCS}$ is the longest common subsequence length;
we set $C_{pt}=1$ when both slot sequences are empty.
It measures slot-type and order correspondence, normalized by sequence
length. The combined score is
\begin{equation}
  \operatorname{sim}(a_p,a_t)
  =\alpha L_{pt}+\beta E_{pt}+\gamma C_{pt},
  \label{eq:pareo-similarity}
\end{equation}
with nonnegative weights satisfying $\alpha+\beta+\gamma=1$.

Starting from singleton forms, complete-link clustering repeatedly merges
the pair of clusters with the largest minimum cross-cluster similarity,
provided that this minimum is at least $\delta$. This prevents chaining:
every pair within a family meets the threshold. The resulting families
$\{\mathcal C_j\}_{j=1}^{J}$ retain their forms, occurrences, and bindings.

While $\delta$ determines family membership, $K$ controls how many families
receive compression policies. We rank families by source-token mass:
equally frequent expressions can cover different amounts of text.
Let $P_{i,n}$ be the source-token positions covered
by candidate $s_{i,n}$ under the target tokenizer.  A form $a$ has mass
$w(a)=\sum_i|\bigcup_{n:\,a_{i,n}=a}P_{i,n}|$.
The mass of a family $\mathcal C_j$ is
\begin{equation}
  W_j=\sum_i\left|\bigcup_{n:\,a_{i,n}\in\mathcal C_j}P_{i,n}\right|.
  \label{eq:pareo-family-mass}
\end{equation}
Unions count overlapping windows only once, both within families and
when measuring coverage across selected families. Selecting the $K$
largest masses, indexed by $\mathcal J_K$, directs shared compression
toward families covering more source tokens.

\subsection{Template Compression}
\label{sec:method-compression}

For each selected family $j\in\mathcal J_K$, we choose a representative
span $s_j^\star$ from the template with the largest source-token mass,
$a_j^\star=\arg\max_{a\in\mathcal C_j}w(a)$. The compressor produces
one shared policy $\pi_j$ for the family.

\noindent\textbf{Deletion Compression.}
The deletion operator takes only $s_j^\star$ as input and scores its
$L_j$ elements, treating each word or recognized slot span as one element.
Let $d_j^{(k)}\in\{0,1\}^{L_j}$ mark the $k$
highest-scoring elements, and let $\tilde s_j^{(k)}$ be their rendering in
source order.  For a requested keep ratio $\eta\in(0,1]$, the shared mask
is chosen by the nearest attainable rendered-token budget:
\begin{equation}
  k_j^\star=\arg\min_{0\leq k\leq L_j}
  \left|\ell(\tilde s_j^{(k)})-\eta\ell(s_j^\star)\right|.
  \label{eq:pareo-deletion-budget}
\end{equation}
The policy stores $d_j=d_j^{(k_j^\star)}$ together with the representative
positions needed to transfer it.  The budget is evaluated using the
target tokenizer and fixed at the representative level.

\noindent\textbf{Rewrite Compression.}
The rewrite teacher receives the representative span along with $m_j$ concrete
occurrences and their surrounding context as examples. Its policy
$\pi_j=\{\tilde a_{j,v}\}_{v=1}^{V_j}$ contains $V_j$ candidate rewrite
templates whose slots take their values from the current occurrence
during application.

\subsection{Template Application}
\label{sec:method-application}

\begingroup
\definecolor{tokenmatchbg}{RGB}{232,241,249}
\definecolor{accmatchbg}{RGB}{234,243,237}
\definecolor{deltapos}{RGB}{25,112,83}
\definecolor{deltaneg}{RGB}{170,58,64}
\newcommand{\metricdelta}[2]{\hspace{0.7pt}{\rmfamily\fontsize{6.5}{7}\bfseries\selectfont\color{#1}#2}}
\newcommand{\blockrule}{\midrule[0.35pt]}
\newcommand{\datasetseparator}{%
\arrayrulecolor{black!30}%
\cmidrule[0.25pt]{1-11}%
\arrayrulecolor{black}}
\newcommand{\tokenmatchbglabel}{Token-match}
\newcommand{\accmatchbglabel}{Acc-match}
\newcommand{\panel}[2]{%
\multicolumn{11}{@{}>{\columncolor{#1}[0pt][0pt]}l@{}}{%
\rule[-2pt]{0pt}{10pt}\hspace{3pt}%
\makebox[\dimexpr\textwidth-6pt\relax][l]{\textbf{#2}\hfill\csname #1label\endcsname}\hspace{3pt}}\\[1pt]}
\newcommand{\tablehead}[1]{%
\toprule
\multirow[c]{2}{*}[-0.2ex]{\textbf{Dataset}} &
\multirow[c]{2}{*}[-0.2ex]{\textbf{Op.}} &
\multicolumn{5}{c}{\textbf{Per-example}} &
\multicolumn{4}{c}{\textbf{CRAFT (Cluster)}}\\
\cmidrule(lr){3-7}\cmidrule(l){8-11}
 & & Lv.
 & #1 & $B\downarrow$ & $T_{\rm in}\downarrow$ & $T_{\rm out}\downarrow$
 & #1 & $B\downarrow$ & $T_{\rm in}\downarrow$ & $T_{\rm out}\downarrow$\\
\midrule}
\newcommand{\matchedtablestyle}{%
\fontsize{8.5}{10}\selectfont
\renewcommand{\arraystretch}{1.00}
\setlength{\tabcolsep}{3pt}
\setlength{\aboverulesep}{0.12ex}
\setlength{\belowrulesep}{0.12ex}
\setlength{\cmidrulesep}{0.12ex}}
\begin{table*}[t]
\centering
\captionsetup{font=small}
\matchedtablestyle
\begin{tabularx}{\textwidth}{@{}llc rrrr
!{\hspace{6pt}}>{\hsize=1.1\hsize\linewidth=\hsize\raggedleft\arraybackslash}X
*{3}{>{\hsize=.9666667\hsize\linewidth=\hsize\raggedleft\arraybackslash}X}@{}}
\tablehead{Acc.$\uparrow$}
\panel{tokenmatchbg}{Qwen2.5-7B}
 & Deletion & 0.5 & 84.46 & 0.962 & 1.642 & 0 & 85.14\metricdelta{deltapos}{+0.68} & 0.058\metricdelta{deltapos}{\textminus{}93.98\%} & 0.099\metricdelta{deltapos}{\textminus{}93.98\%} & 0\metricdelta{black!55}{\textendash{}}\\
\multirow{-2}{*}{GSM8K} & Rewrite & 0.9 & 91.05 & 12.040 & 4.434 & 4.028 & 89.84\metricdelta{deltaneg}{\textminus{}1.21} & 1.794\metricdelta{deltapos}{\textminus{}85.10\%} & 2.258\metricdelta{deltapos}{\textminus{}49.09\%} & 0.201\metricdelta{deltapos}{\textminus{}95.01\%}\\
\datasetseparator
 & Deletion & 0.5 & 56.60 & 1.076 & 3.155 & 0 & 59.80\metricdelta{deltapos}{+3.20} & 0.056\metricdelta{deltapos}{\textminus{}94.83\%} & 0.163\metricdelta{deltapos}{\textminus{}94.83\%} & 0\metricdelta{black!55}{\textendash{}}\\
\multirow{-2}{*}{MATH-500} & Rewrite & 0.9 & 69.00 & 11.845 & 5.252 & 7.367 & 65.40\metricdelta{deltaneg}{\textminus{}3.60} & 1.087\metricdelta{deltapos}{\textminus{}90.83\%} & 2.490\metricdelta{deltapos}{\textminus{}52.59\%} & 0.174\metricdelta{deltapos}{\textminus{}97.64\%}\\
\datasetseparator
 & Deletion & 0.5 & 92.50 & 0.877 & 2.605 & 0 & 91.30\metricdelta{deltaneg}{\textminus{}1.20} & 0.059\metricdelta{deltapos}{\textminus{}93.23\%} & 0.176\metricdelta{deltapos}{\textminus{}93.23\%} & 0\metricdelta{black!55}{\textendash{}}\\
\multirow{-2}{*}{TabMWP} & Rewrite & 0.6 & 96.20 & 18.188 & 11.966 & 10.514 & 95.90\metricdelta{deltaneg}{\textminus{}0.30} & 0.853\metricdelta{deltapos}{\textminus{}95.31\%} & 1.491\metricdelta{deltapos}{\textminus{}87.54\%} & 0.261\metricdelta{deltapos}{\textminus{}97.52\%}\\
\datasetseparator
 & Deletion & 0.7 & 47.90 & 1.056 & 5.043 & 0 & 43.95\metricdelta{deltaneg}{\textminus{}3.95} & 0.059\metricdelta{deltapos}{\textminus{}94.45\%} & 0.280\metricdelta{deltapos}{\textminus{}94.45\%} & 0\metricdelta{black!55}{\textendash{}}\\
\multirow{-2}{*}{LogicNLI} & Rewrite & 0.7 & 50.00 & 8.561 & 10.253 & 7.658 & 45.70\metricdelta{deltaneg}{\textminus{}4.30} & 0.796\metricdelta{deltapos}{\textminus{}90.71\%} & 2.027\metricdelta{deltapos}{\textminus{}80.23\%} & 0.443\metricdelta{deltapos}{\textminus{}94.21\%}\\
\blockrule
\panel{tokenmatchbg}{Qwen2.5-14B}
 & Deletion & 0.6 & 91.96 & 0.922 & 1.616 & 0 & 89.76\metricdelta{deltaneg}{\textminus{}2.20} & 0.105\metricdelta{deltapos}{\textminus{}88.58\%} & 0.185\metricdelta{deltapos}{\textminus{}88.58\%} & 0\metricdelta{black!55}{\textendash{}}\\
\multirow{-2}{*}{GSM8K} & Rewrite & 0.9 & 93.25 & 11.809 & 4.492 & 4.055 & 93.86\metricdelta{deltapos}{+0.61} & 1.720\metricdelta{deltapos}{\textminus{}85.44\%} & 2.214\metricdelta{deltapos}{\textminus{}50.71\%} & 0.200\metricdelta{deltapos}{\textminus{}95.06\%}\\
\datasetseparator
 & Deletion & 0.6 & 69.80 & 0.919 & 2.825 & 0 & 68.60\metricdelta{deltaneg}{\textminus{}1.20} & 0.086\metricdelta{deltapos}{\textminus{}90.67\%} & 0.264\metricdelta{deltapos}{\textminus{}90.67\%} & 0\metricdelta{black!55}{\textendash{}}\\
\multirow{-2}{*}{MATH-500} & Rewrite & 0.9 & 73.20 & 11.487 & 5.486 & 7.460 & 74.20\metricdelta{deltapos}{+1.00} & 1.024\metricdelta{deltapos}{\textminus{}91.09\%} & 2.458\metricdelta{deltapos}{\textminus{}55.19\%} & 0.173\metricdelta{deltapos}{\textminus{}97.69\%}\\
\datasetseparator
 & Deletion & 0.5 & 97.60 & 0.931 & 3.821 & 0 & 95.60\metricdelta{deltaneg}{\textminus{}2.00} & 0.047\metricdelta{deltapos}{\textminus{}94.95\%} & 0.193\metricdelta{deltapos}{\textminus{}94.95\%} & 0\metricdelta{black!55}{\textendash{}}\\
\multirow{-2}{*}{TabMWP} & Rewrite & 0.6 & 99.20 & 15.947 & 13.570 & 12.975 & 98.60\metricdelta{deltaneg}{\textminus{}0.60} & 0.635\metricdelta{deltapos}{\textminus{}96.02\%} & 1.574\metricdelta{deltapos}{\textminus{}88.40\%} & 0.258\metricdelta{deltapos}{\textminus{}98.01\%}\\
\datasetseparator
LogicNLI & Deletion & 0.6 & 51.20 & 0.981 & 5.915 & 0 & 53.35\metricdelta{deltapos}{+2.15} & 0.031\metricdelta{deltapos}{\textminus{}96.82\%} & 0.188\metricdelta{deltapos}{\textminus{}96.82\%} & 0\metricdelta{black!55}{\textendash{}}\\
\tablehead{Tok.$\downarrow$}
\panel{accmatchbg}{Qwen2.5-7B}
 & Deletion & 0.5 & 137.21 & 0.962 & 1.642 & 0 & 136.86\metricdelta{deltapos}{\textminus{}0.26\%} & 0.058\metricdelta{deltapos}{\textminus{}93.98\%} & 0.099\metricdelta{deltapos}{\textminus{}93.98\%} & 0\metricdelta{black!55}{\textendash{}}\\
\multirow{-2}{*}{GSM8K} & Rewrite & 0.8 & 141.90 & 12.040 & 4.434 & 4.028 & 156.33\metricdelta{deltaneg}{+10.17\%} & 1.794\metricdelta{deltapos}{\textminus{}85.10\%} & 2.258\metricdelta{deltapos}{\textminus{}49.09\%} & 0.201\metricdelta{deltapos}{\textminus{}95.01\%}\\
\datasetseparator
 & Deletion & 0.5 & 386.93 & 1.076 & 3.155 & 0 & 365.79\metricdelta{deltapos}{\textminus{}5.46\%} & 0.056\metricdelta{deltapos}{\textminus{}94.83\%} & 0.163\metricdelta{deltapos}{\textminus{}94.83\%} & 0\metricdelta{black!55}{\textendash{}}\\
\multirow{-2}{*}{MATH-500} & Rewrite & 0.8 & 372.77 & 11.845 & 5.252 & 7.367 & 396.48\metricdelta{deltaneg}{+6.36\%} & 1.087\metricdelta{deltapos}{\textminus{}90.83\%} & 2.490\metricdelta{deltapos}{\textminus{}52.59\%} & 0.174\metricdelta{deltapos}{\textminus{}97.64\%}\\
\datasetseparator
LogicNLI & Deletion & 0.6 & 597.64 & 1.056 & 5.043 & 0 & 384.82\metricdelta{deltapos}{\textminus{}35.61\%} & 0.059\metricdelta{deltapos}{\textminus{}94.45\%} & 0.280\metricdelta{deltapos}{\textminus{}94.45\%} & 0\metricdelta{black!55}{\textendash{}}\\
\blockrule
\panel{accmatchbg}{Qwen2.5-14B}
GSM8K & Rewrite & 0.7 & 121.99 & 11.809 & 4.492 & 4.055 & 162.04\metricdelta{deltaneg}{+32.83\%} & 1.720\metricdelta{deltapos}{\textminus{}85.44\%} & 2.214\metricdelta{deltapos}{\textminus{}50.71\%} & 0.200\metricdelta{deltapos}{\textminus{}95.06\%}\\
\datasetseparator
 & Deletion & 0.5 & 387.57 & 0.919 & 2.825 & 0 & 389.80\metricdelta{deltaneg}{+0.58\%} & 0.086\metricdelta{deltapos}{\textminus{}90.67\%} & 0.264\metricdelta{deltapos}{\textminus{}90.67\%} & 0\metricdelta{black!55}{\textendash{}}\\
\multirow{-2}{*}{MATH-500} & Rewrite & 0.9 & 389.34 & 11.487 & 5.486 & 7.460 & 386.39\metricdelta{deltapos}{\textminus{}0.76\%} & 1.024\metricdelta{deltapos}{\textminus{}91.09\%} & 2.458\metricdelta{deltapos}{\textminus{}55.19\%} & 0.173\metricdelta{deltapos}{\textminus{}97.69\%}\\
\datasetseparator
LogicNLI & Deletion & 0.7 & 621.53 & 0.981 & 5.915 & 0 & 546.95\metricdelta{deltapos}{\textminus{}12.00\%} & 0.031\metricdelta{deltapos}{\textminus{}96.82\%} & 0.188\metricdelta{deltapos}{\textminus{}96.82\%} & 0\metricdelta{black!55}{\textendash{}}\\
\bottomrule
\end{tabularx}
\caption{\textbf{CRAFT versus per-example (PE) compression on Qwen2.5-7B and 14B.}
Small bold values show accuracy gains ($A_{\rm CRAFT}-A_{\rm PE}$, pp)
or relative changes in tokens and costs ($100(X_{\rm CRAFT}/X_{\rm PE}-1)$, \%):
green denotes higher accuracy or lower tokens/costs, red the reverse; -- indicates a zero PE denominator.
$B=(T_{\rm in}+4T_{\rm out})/T_{\rm source}$; $T_{\rm in/out}$ are in millions.
Op. = compression operator; Lv. = target retention ratio.}
    \label{tab:combined-matched-results}
\end{table*}
\endgroup

For a candidate occurrence whose abstract form $a_{i,n}$ belongs to a selected
family $\mathcal C_j$, we apply $\pi_j$ using its own source content and
bindings $B_{i,n}$.  This stage reuses the policy without invoking the
compression operator on individual occurrences.

\noindent\textbf{Deletion.}
We align the occurrence's word-and-slot sequence with the
representative using monotonic LCS, treating each slot as a whole element.
Matched elements inherit the shared mask's keep or delete decision.
Unmatched words are omitted, but unmatched or ambiguously aligned slots
are retained intact.  Retained content
is joined in source order with its separators to form $\hat s_{i,n}$.

\noindent\textbf{Rewrite.}
Each $\tilde a_{j,v}$ is rendered by resolving its slot
references against $B_{i,n}$.  Let $\mathcal V_{i,n}$ contain the rendered
variants whose references resolve. We choose
\begin{equation}
  \hat s_{i,n}=\arg\min_{x\in\mathcal V_{i,n}\cup\{s_{i,n}\}}\ell(x),
  \label{eq:pareo-rewrite-application}
\end{equation}
retaining the original candidate when no shorter applicable variant exists.

Since multi-scale candidates can overlap, we select a non-overlapping
set of applicable edits before assembly
(Appendix~E). We construct $\hat c_i$
by replacing each selected candidate's source interval in $c_i$ with
$\hat s_{i,n}$ in source order, leaving text outside the selected intervals
unchanged.

\subsection{Supervised Fine-Tuning}
\label{sec:method-training}

We fine-tune on $\hat{\mathcal D}=\{(q_i,\hat c_i,y_i)\}_{i=1}^{N}$
in the model's native chat format. Assistant targets concatenate the
compact trace, answer delimiter, and original answer; standard
autoregressive loss is evaluated only on assistant-response tokens.

\section{Experiments}
\label{sec:experiments}

\begingroup
\definecolor{tokenmatchbg}{RGB}{232,241,249}
\definecolor{accmatchbg}{RGB}{234,243,237}
\definecolor{deltapos}{RGB}{25,112,83}
\definecolor{deltaneg}{RGB}{170,58,64}
\newcommand{\metricdelta}[2]{\hspace{0.7pt}{\rmfamily\fontsize{6.5}{7}\bfseries\selectfont\color{#1}#2}}
\newcommand{\blockrule}{\midrule[0.35pt]}
\newcommand{\datasetseparator}{%
\arrayrulecolor{black!30}%
\cmidrule[0.25pt]{1-11}%
\arrayrulecolor{black}}
\newcommand{\tokenmatchbglabel}{Token-match}
\newcommand{\accmatchbglabel}{Acc-match}
\newcommand{\panel}[2]{%
\multicolumn{11}{@{}>{\columncolor{#1}[0pt][0pt]}l@{}}{%
\rule[-2pt]{0pt}{10pt}\hspace{3pt}%
\makebox[\dimexpr\textwidth-6pt\relax][l]{\textbf{#2}\hfill\csname #1label\endcsname}\hspace{3pt}}\\[1pt]}
\newcommand{\tablehead}[1]{%
\toprule
\multirow[c]{2}{*}[-0.2ex]{\textbf{Dataset}} &
\multirow[c]{2}{*}[-0.2ex]{\textbf{Op.}} &
\multicolumn{5}{c}{\textbf{Per-example}} &
\multicolumn{4}{c}{\textbf{CRAFT (Cluster)}}\\
\cmidrule(lr){3-7}\cmidrule(l){8-11}
 & & Lv.
 & #1 & $B\downarrow$ & $T_{\rm in}\downarrow$ & $T_{\rm out}\downarrow$
 & #1 & $B\downarrow$ & $T_{\rm in}\downarrow$ & $T_{\rm out}\downarrow$\\
\midrule}
\newcommand{\matchedtablestyle}{%
\fontsize{8.5}{10}\selectfont
\renewcommand{\arraystretch}{1.00}
\setlength{\tabcolsep}{3pt}
\setlength{\aboverulesep}{0.12ex}
\setlength{\belowrulesep}{0.12ex}
\setlength{\cmidrulesep}{0.12ex}}
\begin{table*}[t]
\centering
\captionsetup{font=small}
\matchedtablestyle
\begin{tabularx}{\textwidth}{@{}llc rrrr
!{\hspace{6pt}}>{\hsize=1.1\hsize\linewidth=\hsize\raggedleft\arraybackslash}X
*{3}{>{\hsize=.9666667\hsize\linewidth=\hsize\raggedleft\arraybackslash}X}@{}}
\tablehead{Acc.$\uparrow$}
\panel{tokenmatchbg}{Qwen2.5-3B}
 & Deletion & 0.7 & 79.15 & 0.922 & 1.642 & 0 & 75.28\metricdelta{deltaneg}{\textminus{}3.87} & 0.106\metricdelta{deltapos}{\textminus{}88.51\%} & 0.189\metricdelta{deltapos}{\textminus{}88.51\%} & 0\metricdelta{black!55}{\textendash{}}\\
\multirow{-2}{*}{GSM8K} & Rewrite & 0.9 & 84.38 & 12.252 & 4.414 & 4.348 & 80.97\metricdelta{deltaneg}{\textminus{}3.41} & 1.689\metricdelta{deltapos}{\textminus{}86.21\%} & 2.211\metricdelta{deltapos}{\textminus{}49.90\%} & 0.199\metricdelta{deltapos}{\textminus{}95.43\%}\\
\datasetseparator
MATH-500 & Deletion & 0.5 & 46.80 & 0.915 & 2.388 & 0 & 45.20\metricdelta{deltaneg}{\textminus{}1.60} & 0.107\metricdelta{deltapos}{\textminus{}88.29\%} & 0.280\metricdelta{deltapos}{\textminus{}88.29\%} & 0\metricdelta{black!55}{\textendash{}}\\
\blockrule
\panel{tokenmatchbg}{LLaMA-3.1-8B}
 & Deletion & 0.7 & 80.67 & 0.997 & 1.328 & 0 & 79.08\metricdelta{deltaneg}{\textminus{}1.59} & 0.177\metricdelta{deltapos}{\textminus{}82.24\%} & 0.236\metricdelta{deltapos}{\textminus{}82.24\%} & 0\metricdelta{black!55}{\textendash{}}\\
\multirow{-2}{*}{GSM8K} & Rewrite & 0.9 & 86.05 & 16.646 & 4.019 & 4.538 & 84.46\metricdelta{deltaneg}{\textminus{}1.59} & 2.432\metricdelta{deltapos}{\textminus{}85.39\%} & 2.421\metricdelta{deltapos}{\textminus{}39.77\%} & 0.205\metricdelta{deltapos}{\textminus{}95.49\%}\\
\tablehead{Tok.$\downarrow$}
\panel{accmatchbg}{Qwen2.5-3B}
 & Deletion & 0.5 & 145.05 & 0.922 & 1.642 & 0 & 167.21\metricdelta{deltaneg}{+15.28\%} & 0.106\metricdelta{deltapos}{\textminus{}88.51\%} & 0.189\metricdelta{deltapos}{\textminus{}88.51\%} & 0\metricdelta{black!55}{\textendash{}}\\
\multirow{-2}{*}{GSM8K} & Rewrite & 0.5 & 95.68 & 12.252 & 4.414 & 4.348 & 179.19\metricdelta{deltaneg}{+87.28\%} & 1.689\metricdelta{deltapos}{\textminus{}86.21\%} & 2.211\metricdelta{deltapos}{\textminus{}49.90\%} & 0.199\metricdelta{deltapos}{\textminus{}95.43\%}\\
\datasetseparator
MATH-500 & Rewrite & 0.6 & 288.75 & 11.769 & 4.674 & 6.513 & 453.90\metricdelta{deltaneg}{+57.19\%} & 1.228\metricdelta{deltapos}{\textminus{}89.57\%} & 2.513\metricdelta{deltapos}{\textminus{}46.23\%} & 0.173\metricdelta{deltapos}{\textminus{}97.34\%}\\
\blockrule
\panel{accmatchbg}{LLaMA-3.1-8B}
 & Deletion & 0.6 & 132.24 & 0.997 & 1.328 & 0 & 140.86\metricdelta{deltaneg}{+6.52\%} & 0.177\metricdelta{deltapos}{\textminus{}82.24\%} & 0.236\metricdelta{deltapos}{\textminus{}82.24\%} & 0\metricdelta{black!55}{\textendash{}}\\
\multirow{-2}{*}{GSM8K} & Rewrite & 0.7 & 147.20 & 16.646 & 4.019 & 4.538 & 187.84\metricdelta{deltaneg}{+27.61\%} & 2.432\metricdelta{deltapos}{\textminus{}85.39\%} & 2.421\metricdelta{deltapos}{\textminus{}39.77\%} & 0.205\metricdelta{deltapos}{\textminus{}95.49\%}\\
\datasetseparator
 & Deletion & 0.6 & 389.93 & 0.972 & 1.835 & 0 & 406.31\metricdelta{deltaneg}{+4.20\%} & 0.215\metricdelta{deltapos}{\textminus{}77.85\%} & 0.407\metricdelta{deltapos}{\textminus{}77.85\%} & 0\metricdelta{black!55}{\textendash{}}\\
\multirow{-2}{*}{MATH-500} & Rewrite & 0.7 & 313.17 & 13.253 & 3.959 & 5.267 & 423.02\metricdelta{deltaneg}{+35.08\%} & 1.849\metricdelta{deltapos}{\textminus{}86.04\%} & 2.779\metricdelta{deltapos}{\textminus{}29.81\%} & 0.178\metricdelta{deltapos}{\textminus{}96.61\%}\\
\bottomrule
\end{tabularx}
\caption{\textbf{Additional comparisons on a smaller model (Qwen2.5-3B) and a different model family (LLaMA-3.1-8B).}
See the caption of Table~\ref{tab:combined-matched-results} for notation, abbreviations, and color conventions.}
    \label{tab:additional-backbone-results}
\end{table*}
\endgroup

\subsection{Experimental Setup}
\label{sec:experimental-setup}

\paragraph{Models and data.}
We use Qwen2.5-Instruct (3B, 7B, and 14B)
\citep{qwen-team-2025-qwen25} and LLaMA-3.1-8B-Instruct
\citep{dubey-etal-2024-llama3} to generate CoT rollouts on the GSM8K
\citep{cobbe-etal-2021-gsm8k} and MATH \citep{hendrycks-etal-2021-math}
training sets. For TabMWP \citep{lu-etal-2023-tabmwp} and LogicNLI
\citep{tian-etal-2021-logicnli}, we generate training CoT rollouts using
Qwen2.5-Instruct 7B and 14B only.
CRAFT, TokenSkip, and C3oT compress the same correctness-filtered rollouts to
fine-tune the corresponding backbone.  CRAFT constructs a separate policy set
for each model--dataset pair.
We evaluate on GSM8K test, MATH-500, TabMWP test1k, and the
2,000-question LogicNLI test set.

Implementation, training, and hardware details are provided in
Appendix~B.

\paragraph{Baselines and operating points.}
\begingroup
\definecolor{zerocostbg}{RGB}{232,241,249}
\definecolor{crosssamplebg}{RGB}{234,243,237}
\definecolor{modelbandbg}{RGB}{244,245,246}
\newcommand{\lowresourcemodel}[1]{%
\multicolumn{10}{@{}>{\columncolor{modelbandbg}[0pt][0pt]}l@{}}{%
\rule[-2pt]{0pt}{10pt}\hspace{3pt}%
\makebox[\dimexpr\textwidth-6pt\relax][l]{\textbf{#1}}\hspace{3pt}}\\[1pt]}
\newcommand{\lowresourcerule}{%
\arrayrulecolor{black!30}\midrule[0.25pt]\arrayrulecolor{black}}
\begin{table*}[t]
\centering
\captionsetup{font=small}
\fontsize{8.5}{10}\selectfont
\renewcommand{\arraystretch}{1.00}
\setlength{\tabcolsep}{3pt}
\setlength{\aboverulesep}{0.12ex}
\setlength{\belowrulesep}{0.12ex}
\setlength{\cmidrulesep}{0.12ex}
\begin{tabularx}{\textwidth}{lc *{8}{>{\raggedleft\arraybackslash}X}}
\toprule
\multirow[c]{2}{*}[-0.2ex]{\textbf{Method}} &
\multirow[c]{2}{*}[-0.2ex]{\textbf{Lv.}} &
\multicolumn{2}{c}{\textbf{GSM8K}} &
\multicolumn{2}{c}{\textbf{MATH-500}} &
\multicolumn{2}{c}{\textbf{TabMWP}} &
\multicolumn{2}{c}{\textbf{LogicNLI}}\\
\cmidrule{3-10}
 & & Acc.$\uparrow$ & Tok.$\downarrow$ & Acc.$\uparrow$ & Tok.$\downarrow$
 & Acc.$\uparrow$ & Tok.$\downarrow$ & Acc.$\uparrow$ & Tok.$\downarrow$\\
\midrule
\lowresourcemodel{Qwen2.5-7B}
Full CoT & Fixed & 91.13 & 252.90 & 68.00 & 566.49 & 93.40 & 142.53 & 46.15 & 677.30\\
Full CoT SFT & Fixed & 91.43 & 254.50 & 70.00 & 565.48 & 95.20 & 142.95 & 45.65 & 683.10\\
No CoT & Fixed & 24.72 & 0.00 & 27.40 & 0.00 & 65.00 & 0.00 & 27.30 & 0.00\\
\lowresourcerule
\rowcolor{zerocostbg} & 0.5 & 90.60 & 199.98 & 65.00 & 464.10 & 94.00 & 119.72 & 42.75 & 606.47\\
\rowcolor{zerocostbg} & 0.7 & 90.14 & 211.51 & 65.60 & 492.15 & 94.40 & 131.50 & 43.50 & 637.04\\
\rowcolor{zerocostbg}\multirow{-3}{*}{LC Prompt} & 0.9 & 90.98 & 224.33 & 65.40 & 526.38 & 94.30 & 140.18 & 43.10 & 653.19\\
\rowcolor{zerocostbg} & 0.5 & 18.95 & 126.20 & 26.80 & 296.54 & 65.50 & 71.02 & 32.15 & 338.41\\
\rowcolor{zerocostbg} & 0.7 & 34.87 & 176.58 & 35.80 & 415.03 & 78.80 & 99.34 & 37.80 & 473.66\\
\rowcolor{zerocostbg}\multirow{-3}{*}{Truncation} & 0.9 & 49.43 & 227.15 & 63.20 & 533.74 & 85.90 & 127.84 & 42.10 & 609.12\\
\lowresourcerule
\rowcolor{crosssamplebg}Supertoken\textsuperscript{\dag} & Fixed & 88.93 & 208.12 & 66.60 & 507.78 & 92.70 & 97.56 & 48.50 & 544.87\\
\lowresourcerule
\textbf{CRAFT (Rewrite)} & Fixed & 89.84 & 156.33 & 65.40 & 396.48 & 95.90 & 81.16 & 45.70 & 349.18\\
\midrule[0.35pt]
\lowresourcemodel{Qwen2.5-14B}
Full CoT & Fixed & 92.72 & 257.19 & 71.60 & 580.71 & 98.40 & 186.55 & 51.10 & 752.29\\
Full CoT SFT & Fixed & 94.31 & 256.52 & 72.80 & 566.76 & 98.80 & 186.52 & 52.10 & 762.41\\
No CoT & Fixed & 34.87 & 0.00 & 31.60 & 0.00 & 73.10 & 0.00 & 37.55 & 0.00\\
\lowresourcerule
\rowcolor{zerocostbg} & 0.5 & 93.03 & 115.87 & 70.60 & 312.22 & 94.20 & 68.00 & 47.30 & 239.29\\
\rowcolor{zerocostbg} & 0.7 & 93.33 & 148.36 & 73.60 & 412.26 & 97.00 & 106.66 & 47.40 & 315.66\\
\rowcolor{zerocostbg}\multirow{-3}{*}{LC Prompt} & 0.9 & 93.56 & 155.62 & 72.80 & 425.71 & 97.10 & 113.51 & 47.80 & 377.19\\
\rowcolor{zerocostbg} & 0.5 & 33.43 & 128.35 & 35.80 & 307.66 & 79.40 & 93.02 & 46.85 & 375.90\\
\rowcolor{zerocostbg} & 0.7 & 52.62 & 179.58 & 46.20 & 430.61 & 86.80 & 130.14 & 47.80 & 526.16\\
\rowcolor{zerocostbg}\multirow{-3}{*}{Truncation} & 0.9 & 74.22 & 231.01 & 66.20 & 553.78 & 94.50 & 167.44 & 51.45 & 676.62\\
\lowresourcerule
\rowcolor{crosssamplebg}Supertoken\textsuperscript{\dag} & Fixed & 92.72 & 214.74 & 71.20 & 505.71 & 97.50 & 141.06 & 52.75 & 663.74\\
\lowresourcerule
\textbf{CRAFT (Rewrite)} & Fixed & 93.86 & 162.04 & 74.20 & 386.39 & 98.60 & 96.63 & 49.90 & 508.37\\
\bottomrule
\end{tabularx}
\caption{\textbf{Accuracy--length tradeoffs for CRAFT Rewrite versus low-resource baselines on Qwen2.5-7B and 14B.}
Acc. = accuracy (\%); Tok. = mean reasoning length in tokens.
Blue: zero-compressor-cost methods; green: cross-sample encoding.
Lv. = target retention ratio; Fixed denotes a single operating point.
\textsuperscript{\dag}Supertoken uses Top-1,500 rules and partial unfreezing;
each supertoken counts as one token.}
\label{tab:low-resource-comparison}
\end{table*}
\endgroup

\textsc{Full CoT} generates complete reasoning traces, while
\textsc{No-CoT} answers directly without intermediate reasoning; both use
instruction models without SFT. We compare per-example compression with
CRAFT's cluster-level compression using the same operator. Per-example
deletion reproduces \textsc{TokenSkip} \citep{xia-etal-2025-tokenskip},
and per-example teacher rewriting reproduces \textsc{C3oT}
\citep{kang-etal-2024-c3ot}.
For cross-sample encoding, Supertoken uses the top 1,500 merge rules
with partial unfreezing on Qwen2.5-7B/14B; Appendix~H
also reports 10,000 rules with the same training configuration.

\paragraph{Metrics and matching protocol.}
We report accuracy $A$ (\%) and mean CoT length $L$ in backbone tokens,
excluding prompts, answer markers, and answers, with a shared answer
verifier. Generation limits are 512 (GSM8K), 1,024 (MATH-500), and
2,048 tokens (TabMWP/LogicNLI). Compressor input and
output totals, $T_{\rm in}$ and $T_{\rm out}$, are reported in millions.
Performance--budget comparisons use
\begin{equation}
  B=\frac{T_{\rm in}+\lambda T_{\rm out}}{T_{\rm source}},
  \label{eq:normalized-compressor-budget}
\end{equation}
where $T_{\rm source}$ is the original training CoT token total, shared
within each model--dataset pair. We set $\lambda=4$, the teacher model's
official output-to-input token price ratio. Reasoning compression rate is
$100(1-L/L_{\rm full})$, with $L_{\rm full}$ the mean Full CoT length.

Within each model--dataset--operator comparison, Acc-match selects the
per-example setting closest in accuracy but below CRAFT to compare
compression rate and budget; Token-match selects the setting closest
in length but longer than CRAFT to compare accuracy and budget.
Formal definitions and tie-breaking rules are in
Appendix~C.

\begin{figure*}[t]
  \centering
  \includegraphics[width=\textwidth]{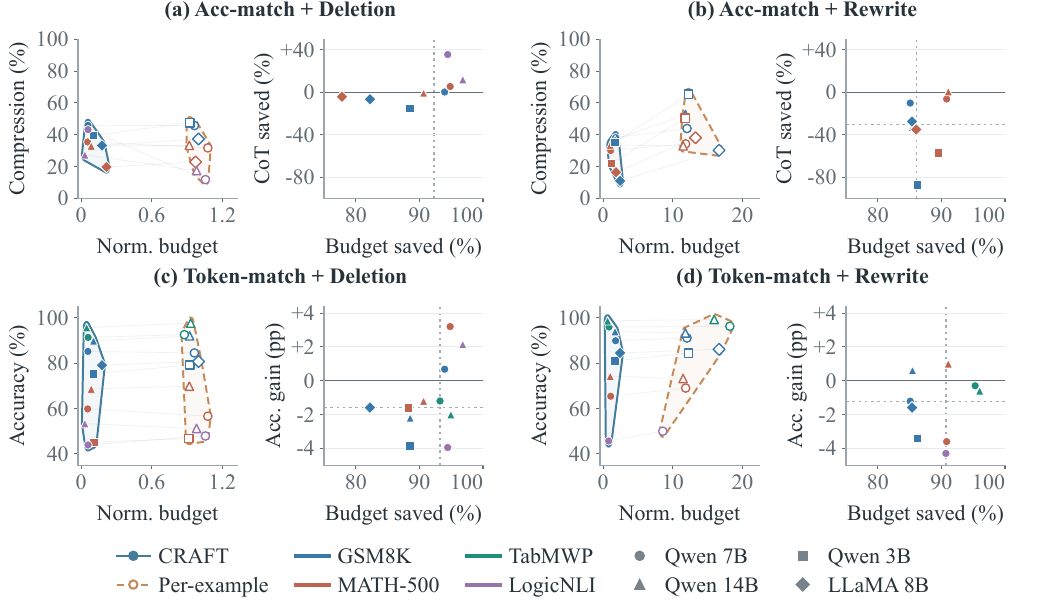}
  \caption{Compressor budget and downstream performance under
  (a,b) accuracy matching and (c,d) token matching. Left: absolute
  comparisons; right: paired CRAFT gains over per-example compression.
  Blue solid/orange dashed envelopes denote CRAFT/per-example results;
  dotted lines mark medians. Colors and shapes denote datasets and backbones.}
  \label{fig:matched-budget}
  \label{fig:paired-budget-performance-gains}
\end{figure*}

\section{Results}
\label{sec:results}

\subsection{Overall}
\label{sec:main-results}

\paragraph{Shared edits sharply reduce compression cost.}
CRAFT uses 77.8--96.8\% less weighted compressor budget than per-example
compression across 36 matched comparisons
(Tables~\ref{tab:combined-matched-results} and~\ref{tab:additional-backbone-results};
Figure~\ref{fig:matched-budget}).
The 16 accuracy-matched and 20 token-matched comparisons yield
budget-saving medians of 86.1--93.2\% across the four groups.
These savings come from reusing deletion masks and rewrite policies
across occurrences.

\paragraph{Lower budgets with competitive downstream performance.}
The paired plots in Figure~\ref{fig:matched-budget} quantify the
associated performance changes. Under Token-match, CRAFT Rewrite combines
shorter reasoning with a median budget reduction of 90.7\% and a median
accuracy change of $-1.21$ percentage points.

\paragraph{Stronger tradeoffs than cross-sample encoding.}
Table~\ref{tab:low-resource-comparison} compares CRAFT Rewrite with other
low-resource methods through accuracy and reasoning length.
Against Top-1500 Supertoken, CRAFT produces 16.8--35.9\% shorter reasoning
across all eight model--dataset combinations and higher accuracy in five.
Shared textual edits thus support simultaneous gains
in answer accuracy and generated length over vocabulary encoding.
Appendix~H compares both rule inventories:
increasing Supertoken to 10,000 rules leaves CRAFT ahead in both metrics
in the same five settings.

\paragraph{Accuracy--length gains over zero-cost controls.}
Compared with the most accurate Truncation result among the three retention
levels for each model--dataset combination, CRAFT achieves higher accuracy
and shorter reasoning in seven of eight cases.
Against each setting's highest-accuracy LC Prompt level, CRAFT improves
both metrics in four of eight cases.
Relative to Full CoT on LLaMA-3.1-8B-Instruct, LC Prompt adds
8.7--50.5 reasoning tokens, whereas CRAFT Rewrite saves 23.0 tokens
on GSM8K and 82.6 tokens on MATH-500.
LC Prompt can thus yield negative compression rather than consistently
shortening reasoning across model families (Appendix~P,
Table~14).

\subsection{Ablation: Offline Compression on Held-Out Traces}
\label{sec:textual-redundancy}

\begin{figure}[htbp]
  \centering
  \includegraphics[width=\linewidth]{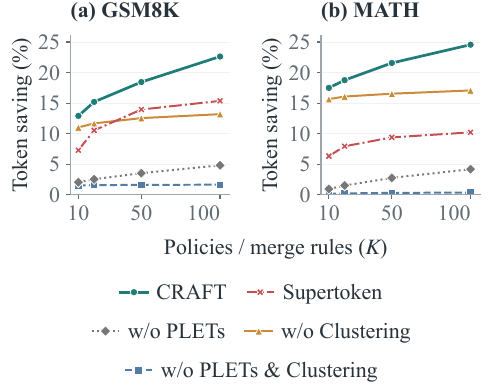}
  \caption{Held-out token saving versus bank size ($K$) for Qwen2.5-7B
  on GSM8K and MATH.}
  \label{fig:heldout-semantic-saving}
\end{figure}

Shared compression policies generalize beyond the traces used to construct
them. Figure~\ref{fig:heldout-semantic-saving} applies banks built from
80\% of the documents to the held-out 20\%, keeping policies and merge
rules fixed. This tests whether decisions made once can compress new
examples. Here, $K$ counts reusable edit policies or merge rules, and
curves track token reduction as the bank grows. Savings are normalized
by held-out Full CoT tokens; edit-based methods count reductions within
matched occurrences, while Supertoken uses the reduction in encoded length.

\paragraph{Abstraction and clustering broaden reuse.}
We ablate PLET abstraction and clustering in the deletion-based pipeline.
CRAFT clusters similar abstract forms; \emph{w/o PLETs} clusters source
text without abstraction; \emph{w/o Clustering} groups identical PLETs;
and \emph{w/o PLETs \& Clustering} groups identical source text.
At $K=100$, \emph{w/o Clustering} saves 13.19\% on GSM8K and 17.07\% on MATH,
against 1.66\% and 0.37\% for \emph{w/o PLETs \& Clustering}.
CRAFT further raises saving to
22.61\% and 24.55\%, respectively.
Abstraction removes instance-specific differences, and clustering
additionally reuses edits across wording variants: both steps extend
compression beyond repeated strings.

\paragraph{A smaller policy bank delivers greater savings.}
CRAFT achieves the highest held-out saving at every tested $K$ on both
datasets. Supertoken encodes merged sequences as single vocabulary tokens;
at $K=100$, CRAFT saves $1.47\times$ and $2.40\times$ as many tokens on
GSM8K and MATH, respectively.
With 50 policies, CRAFT saves 18.44\% and 21.56\%, exceeding every
alternative with 100 policies or merge rules.
Grouping expressions by their shared structure lets each policy apply
to more variations, so a smaller bank can compress more unseen text.

Top-1500 inventories also cover 72.7--98.8\% of eligible training tokens
for Qwen2.5-7B/14B (Appendix~G).

\subsection{Ablation: Downstream Accuracy and Reasoning Length}
\label{sec:plet-clustering-ablation}

PLET abstraction and clustering jointly strengthen compression in both
training targets and generated reasoning
(Table~\ref{tab:plet-clustering-ablation}). With $K=1500$, CRAFT Deletion
reduces training-target tokens by 46.98\% and generated CoT length by
45.88\%. Both \emph{w/o PLETs \& Clustering} and \emph{w/o PLETs} match
Full CoT accuracy (91.13\%) but shorten generated reasoning by just
2.82\% and 3.42\%, respectively.

\begin{table}[!htbp]
\centering
\small
\setlength{\tabcolsep}{3pt}
\renewcommand{\arraystretch}{1.15}
\begin{tabular*}{\columnwidth}{@{\extracolsep{\fill}}lrrrr@{}}
\toprule
Method & Acc.\,$\uparrow$ & Tok.\,$\downarrow$ &
CR\,$\uparrow$ & Target \\
\midrule
Full CoT & 91.13 & 252.90 & 0.00 & 0.00 \\
\midrule
\shortstack[l]{w/o PLETs \&\\Clustering} & 91.13 & 245.77 & 2.82 & 5.92 \\
w/o Clustering & 87.71 & 211.94 & 16.20 & 21.90 \\
w/o PLETs & 91.05 & 244.24 & 3.42 & 7.56 \\
\midrule
\textbf{CRAFT Deletion} & 85.14 & 136.86 & 45.88 & 46.98 \\
\bottomrule
\end{tabular*}
\caption{PLET and clustering ablations on Qwen2.5-7B GSM8K
with $K=1500$. Tok. is mean generated CoT length in tokens; CR is its reduction
relative to Full CoT. Target denotes the percentage reduction in
offline training-target tokens.}
\label{tab:plet-clustering-ablation}
\end{table}

Clustering extends this benefit across wording variants. Relative to
\emph{w/o Clustering}, the full method trades 2.57 percentage points of
accuracy for 35.43\% fewer generated reasoning tokens (211.94 to 136.86),
while increasing training-target savings by 25.08 percentage points.
Together, the two components turn recurring
expressions into compact supervision whose compression benefits carry
through to model generation.

\section{Conclusion}
\label{sec:conclusion}

CRAFT abstracts problem-specific details and clusters wording variants
to make expressions reusable across reasoning traces. Shared edits reduce
weighted compressor budgets by 77.8--96.8\% across 36 matched comparisons
with per-example compression, while maintaining competitive downstream
performance. Shared rewriting also delivers stronger accuracy--length
tradeoffs than cross-sample encoding and zero-cost baselines across
multiple model--dataset settings. The resulting supervision remains
ordinary text, enabling models to learn concise reasoning through
standard supervised fine-tuning without additional inference-time
machinery.

\section*{Limitations}
\label{sec:limitations}

Our experiments cover English mathematical, tabular, and logical reasoning.
The current abstraction uses English language tools and task-relevant
content recognizers; extending it to multilingual, code, or open-ended
reasoning requires further evaluation. Reuse also depends on recurring
expression patterns, so more diverse corpora may offer fewer shared edits.
We build a separate policy bank for each model--dataset pair. The transfer
experiment in Appendix~J evaluates trained models, not the transfer of a policy bank
between corpora.

CRAFT discovers shared expressions by analyzing a corpus of reasoning
traces before compression. It is designed for offline, batch construction
of training supervision rather than immediate compression of individual
requests as they arrive.

CRAFT compresses local expressions through shared deletion and rewriting
edits. Its scope is to shorten how reasoning steps are expressed, rather
than to reorganize the solution steps or search for a shorter solution.

\section*{Ethical Considerations}
\label{sec:ethical-considerations}

This work uses publicly released reasoning benchmarks and model-generated
traces, with no new human-subject data collection. Use of the datasets,
models, and generated supervision should follow their licenses and terms.
Teacher rewriting sends source examples to an external service; applying
it to private data requires authorization and appropriate data-handling
protections.

Shorter reasoning can be harder to audit when compression removes context
or intermediate steps. Answer accuracy alone does not make a compressed
trace a faithful explanation, and shared edits can spread errors across
examples. Applications requiring auditability should retain the original
traces and independently check consequential outputs rather than rely on
compressed explanations alone. We report token usage as a resource metric,
not as a measurement of energy consumption or carbon emissions.

\bibliography{custom,additional_references_20261008,intro_references_20261010}

\end{document}